\documentclass[11pt,letterpaper]{article}
\usepackage{emnlp2017}
\usepackage{times}
\usepackage{latexsym}

\usepackage[pdftex]{graphicx}
\graphicspath{{../pdf/}{./jpeg/}{./images/}}
\DeclareGraphicsExtensions{.pdf,.jpeg,.png}
\usepackage{amssymb}
\usepackage{esvect}
\newcommand{\cev}[1]
{\reflectbox{\ensuremath{\vec{\reflectbox{\ensuremath{#1}}}}}}

\emnlpfinalcopy

\def\emnlppaperid{33}

\title{A Multi-task Approach for Named Entity Recognition in \\ Social Media Data}

\author{Gustavo Aguilar, Suraj Maharjan, 
		A. Pastor L\'opez-Monroy \and Thamar Solorio \\
        Department of Computer Science\\
        University of Houston \\
        Houston, TX 77204-3010 \\
  {\tt \{gaguilaralas, smaharjan2, alopezmonroy, tsolorio\}@uh.edu}}

\date{}

\begin{document}

\maketitle

\begin{abstract}
Named Entity Recognition for social media data is challenging because of its inherent noisiness. In addition to improper grammatical structures, it contains spelling inconsistencies and numerous informal abbreviations. We propose a novel multi-task approach by employing a more general secondary task of Named Entity (NE) segmentation together with the primary task of fine-grained NE categorization. The multi-task neural network architecture learns higher order feature representations from word and character sequences along with basic Part-of-Speech tags and gazetteer information. This neural network acts as a feature extractor to feed a Conditional Random Fields classifier. We were able to obtain the first position in the 3rd Workshop on Noisy User-generated Text (WNUT-2017) with a 41.86\% entity F1-score and a 40.24\% surface F1-score. 
	
\end{abstract}

\section{Introduction}

Named Entity Recognition (NER) aims at identifying different types of entities, such as people names, companies, location, etc., within a given text. This information is useful for higher-level Natural Language Processing (NLP) applications such as information extraction, summarization, and data mining \cite{Chen:2004:CDM:987521.987556, Banko:2007:OIE:1625275.1625705, Aramaki:2009:TMT:1572364.1572390}. Learning Named Entities (NEs) from social media is a challenging task mainly because (i) entities usually represent a small part of limited annotated data which makes the task hard to generalize, and (ii) they do not follow strict rules \cite{Ritter:2011:NER:2145432.2145595, Li:2012:TNE:2348283.2348380}. 

This paper describes a multi-task neural network that aims at generalizing the underneath rules of emerging NEs in user-generated text. In addition to the main category classification task, we employ an auxiliary but related secondary task called NE segmentation (i.e. a binary classification of whether a given token is a NE or not). We use both tasks to jointly train the network. More specifically, the model captures word shapes and some orthographic features at the character level by using a Convolutional Neural Network (CNN). For contextual and syntactical information at the word level, such as word and Part-of-Speech (POS) embeddings, the model implements a Bidirectional Long-Short Term Memory (BLSTM) architecture. Finally, to cover well-known entities, the model uses a dense representation of gazetteers. Once the network is trained, we use it as a feature extractor to feed a Conditional Random Fields (CRF) classifier. The CRF classifier jointly predicts the most likely sequence of labels giving better results than the network itself.

With respect to the participants of the shared task, our approach achieved the best results in both categories: 41.86\% F1-score for entities, and 40.24\% F1-score for surface forms. The data for this shared task is provided by \citet{wnutOrganizers17}.

\section{Related Work}

Traditional NER systems use hand-crafted features, gazetteers and other external resources to perform well \cite{Ratinov:2009:DCM:1596374.1596399}. \citet{luo-EtAl:2015:EMNLP2} obtain state-of-the-art results by relying on heavily hand-crafted features, which are expensive to develop and maintain. Recently, many studies have outperformed traditional NER systems by applying neural network architectures. For instance, \citet{glample2016} use a bidirectional LSTM-CRF architecture. They obtain a state-of-the-art performance without relying on hand-crafted features. \citet{limsopatham2016_wnut_ner}, who achieved the first place on WNUT-2016 shared task, use a BLSTM neural network to leverage orthographic features. We use a similar approach but we employ CNN and BLSTM in parallel instead of forwarding the CNN output to the BLSTM. Nevertheless, our main contribution resides on Multi-Task Learning (MTL) and a combination of POS tags and gazetteers representation to feed the network.
    
    Recently, MTL has gained significant attention. Researchers have tried to correlate the success of MTL with label entropy, regularizers, training data size, and other aspects~\cite{martinezalonso-plank:2017:EACLlong,bingel-sogaard:2017:EACLshort}. For instance, \newcite{collobert2008unified} use a multi-task network for different NLP tasks and show that the multi-task setting improves generality among shared tasks. In this paper, we take advantage of the multi-task setting by adding a more general secondary task, NE segmentation, along with the primary NE categorization task.

\section{Methodology}

This section describes our system\footnote{~\url{https://github.com/tavo91/NER-WNUT17}}
in three parts: feature representation, model description\footnote{~The neural network is implemented using Keras (\url{https://github.com/fchollet/keras}) and Theano as backend (\url{http://deeplearning.net/software/theano/}).}, and sequential inference. 

\subsection{Feature Representation} \label{feature_rep}

We select features to represent the most relevant aspects of the data for the task. The features are divided into three categories: character, word, and lexicons.

\noindent{\bf Character representation}: we use an orthographic encoder similar to that of \citet{limsopatham2016_wnut_ner} to encapsulate capitalization, punctuation, word shape, and other orthographic features. The only difference is that we handle non-ASCII characters. For instance, the sentence \emph{``3rd Workshop !''} becomes \emph{``ncc Cccccccc p''} as we map numbers to `n', letters to `c' (or `C' if capitalized), and punctuation marks to `p'. Non-ASCII characters are mapped to `x'. This encoded representation reduces the sparsity of character features and allows us to focus on word shapes and punctuation patterns. Once we have an encoded word, we represent each character with a 30-dimensional vector \cite{DBLP:journals/corr/MaH16}. We account for a maximum length of 20 characters\footnote{~Different lengths do not improve results} per word, applying post padding on shorter words and truncating longer words.

\noindent{\bf Word representation}: we have two different representations at the word level. The first one uses pre-trained word embeddings trained on 400 million tweets representing each word with 400 dimensions \cite{godin2015multimedia}\footnote{~\url{http://www.fredericgodin.com/software}}. The second one uses Part-of-Speech tags generated by the CMU Twitter POS tagger \cite{owoputi2013improved}. The POS tag embeddings are represented by 100-dimensional vectors. In order to capture contextual information, we account for a context window of 3 tokens on both words and POS tags, where the target token is in the middle of the window.

We randomly initialize both the character features and the POS tag vectors using a uniform distribution in the range \(\left[-\sqrt{\frac{3}{dim}}, +\sqrt{\frac{3}{dim}}\right]\), where \(dim\) is the dimension of the vectors from each feature representation \cite{he2015delving}.

\noindent{\bf Lexical representation}: we use gazetteers provided by \citet{mishra2016_wnut_ner} to help the model improve its precision for well-known entities. For each word we create a binary vector of 6 dimensions (one dimension per class). Each of the vector dimensions is set to one if the word appears in the gazetteers of the related class.

\subsection{Model Description} \label{model}

\begin{figure}
\centering
\includegraphics[width=\linewidth,height=6cm]{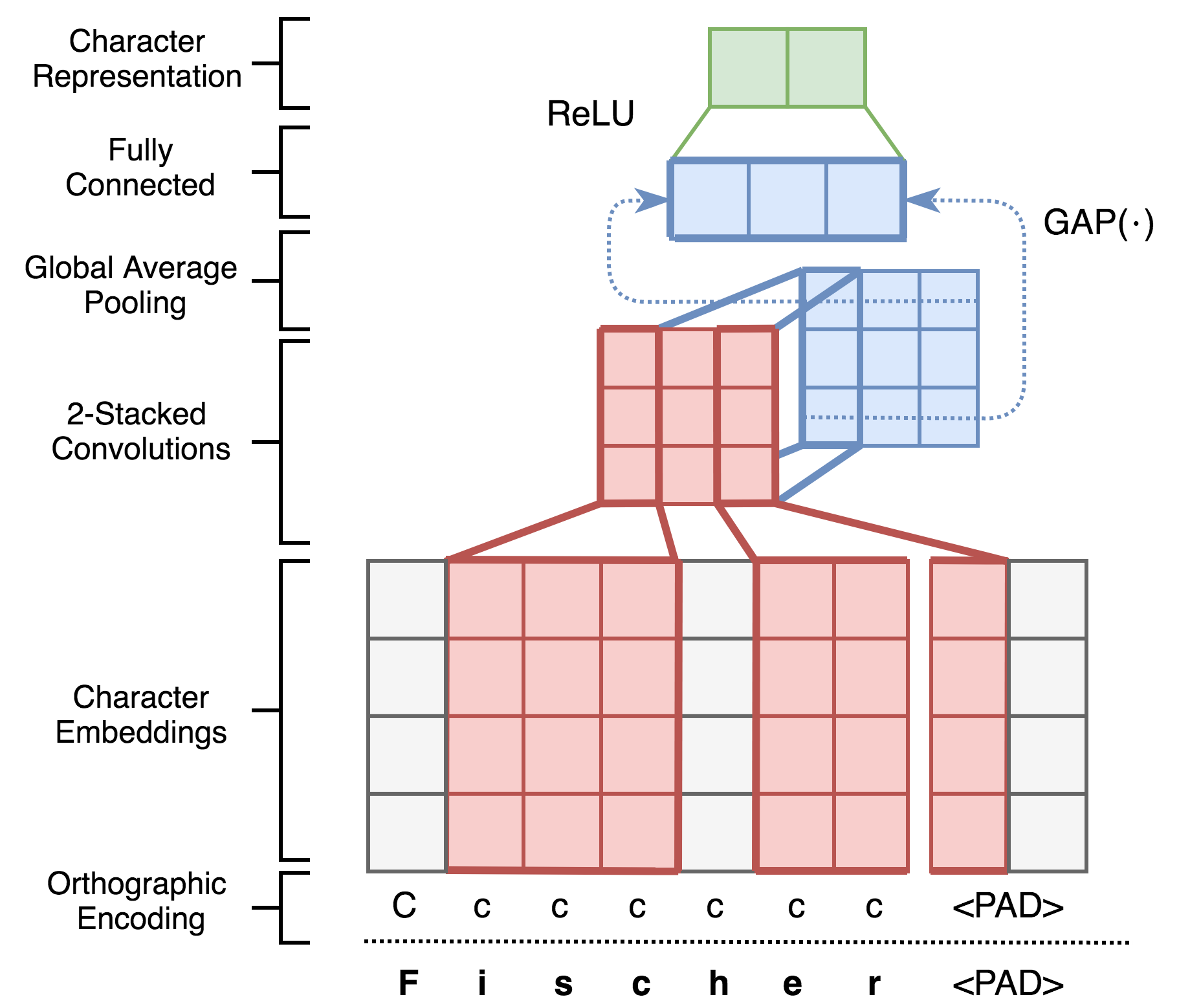}
\caption{ \small Orthographic character-based representation of a word (green) using a CNN with 2-stacked convolutional layers. The first layer takes the input from embeddings (red) while the second layer (blue) takes the input from the first convolutional layer. Global Average Pooling is applied after the second convolutional layer.}
\label{fig:cnn_char}
\end{figure}

\noindent{\bf Character level CNN}: we use a CNN architecture to learn word shapes and some orthographic features at the character level representation (see Figure \ref{fig:cnn_char}). The characters are embedded into a $\mathbb{R}^{d \times l}$ dimensional space, where $d$ is the dimension of the features per character and $l$ is the maximum length of characters per word. Then, we take the character embeddings and apply 2-stacked convolutional layers. Following \citet{DBLP:journals/corr/ZhouKLOT15}, we perform a \textit{global average pooling}\footnote{~\citet{DBLP:journals/corr/ZhouKLOT15} empirically showed that \textit{global average pooling} captured more extensive information from the feature maps than \textit{max pooling}.} instead of the widely used \textit{max pooling} operation. Finally, the result is passed to a fully-connected layer using a Rectifier Linear Unit (ReLU) activation function, which yields the character-based representation of a word. The resulting vector is used as input for the rest of the network.

\begin{figure}
\centering
\includegraphics[width=\linewidth,height=6cm]{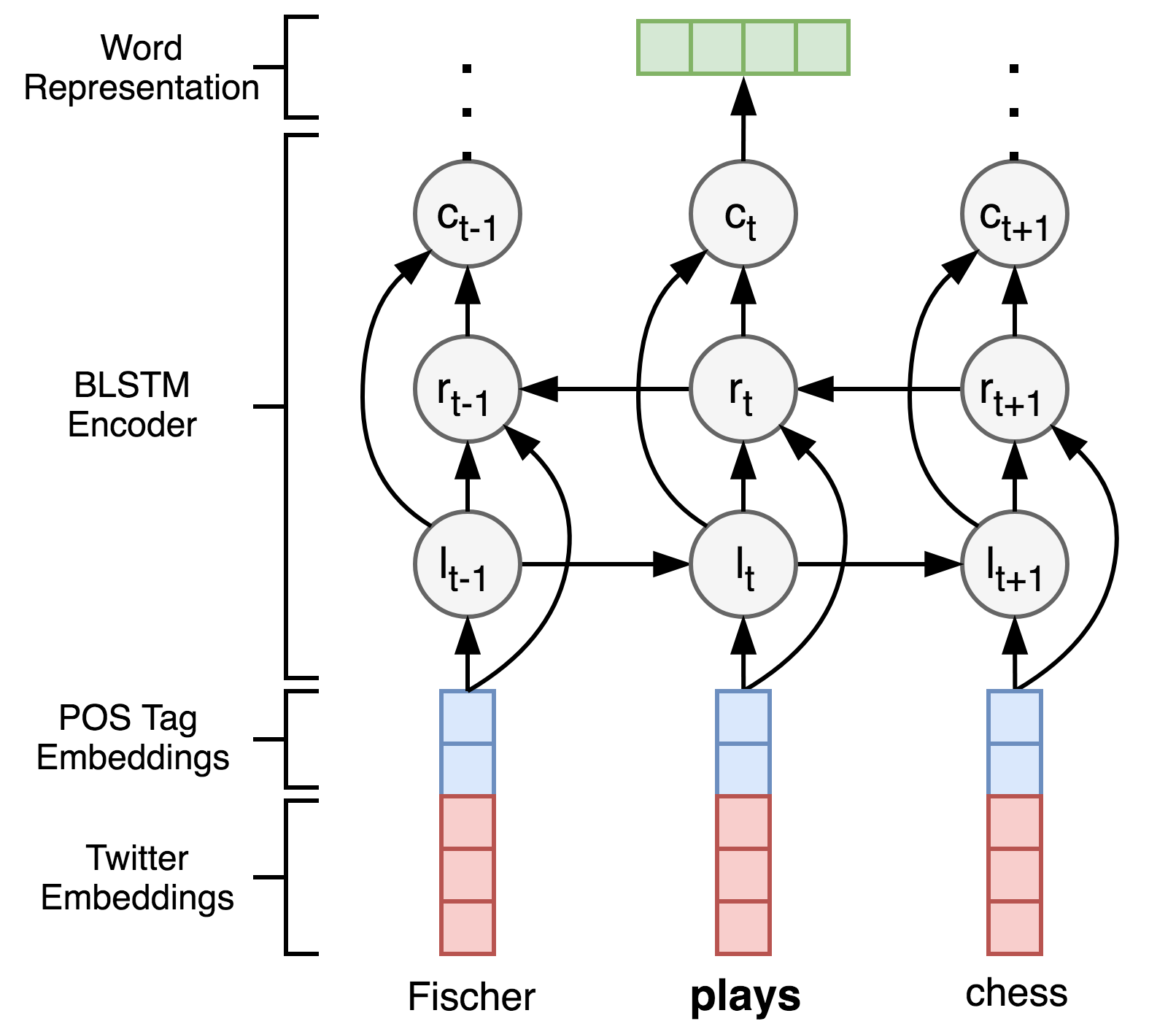}
\caption{ \small Word representation of POS-tag embeddings (blue) and Twitter word embeddings (red) using a BLSTM neural network. }
\label{fig:blstm_word}
\end{figure}

\noindent{\bf Word level BLSTM}: we use a Bidirectional LSTM \cite{DBLP:journals/corr/DyerBLMS15} to learn the contextual information of a sequence of words as described in Figure \ref{fig:blstm_word}. Word embeddings are initialized with pre-trained Twitter word embeddings from a Skip-gram model \cite{godin2015multimedia} using word2vec \cite{mikolov2013word2vec}. Additionally, we use POS tag embeddings, which are randomly initialized using a uniform distribution. The model receives the concatenation of both POS tags and Twitter word embeddings. The BLSTM layer extracts the features from both forward and backward directions and concatenates the resulting vectors from each direction ($[\vec{h};\quad\cev{h}]$). Following \citet{DBLP:journals/corr/MaH16}, we use 100 neurons per direction. The resulting vector is used as input for the rest of the network.

\noindent{\bf Lexicon network}: we take the lexical representation vectors of the input words and feed them into a fully-connected layer. We use 32 neurons on this layer and a ReLU activation function. Then, the resulting vector is used as input for the rest of the network.

\noindent{\bf Multi-task network}: we create a unified model to predict the NE segmentation and NE categorization tasks simultaneously. Typically, the additional task acts as a regularizer to generalize the model~\cite{goodfellow2016deep,collobert2008unified}. The concatenation of character, word and lexical vectors is fed into the NE segmentation and  categorization tasks. We use a single-neuron layer with a sigmoid activation function for the secondary NE segmentation task, whereas for the primary NE categorization task, we employ a 13-neuron\footnote{~Using BIO encoding, each of the 6 classes will have a \textit{begin} and \textit{inside} version (e.g. B-product, I-product).} layer with a softmax activation function. Finally, we add the losses from both tasks and feed the total loss backward during training.

\subsection{Sequential Inference} \label{crf}

\begin{figure}
\includegraphics[width=\linewidth]{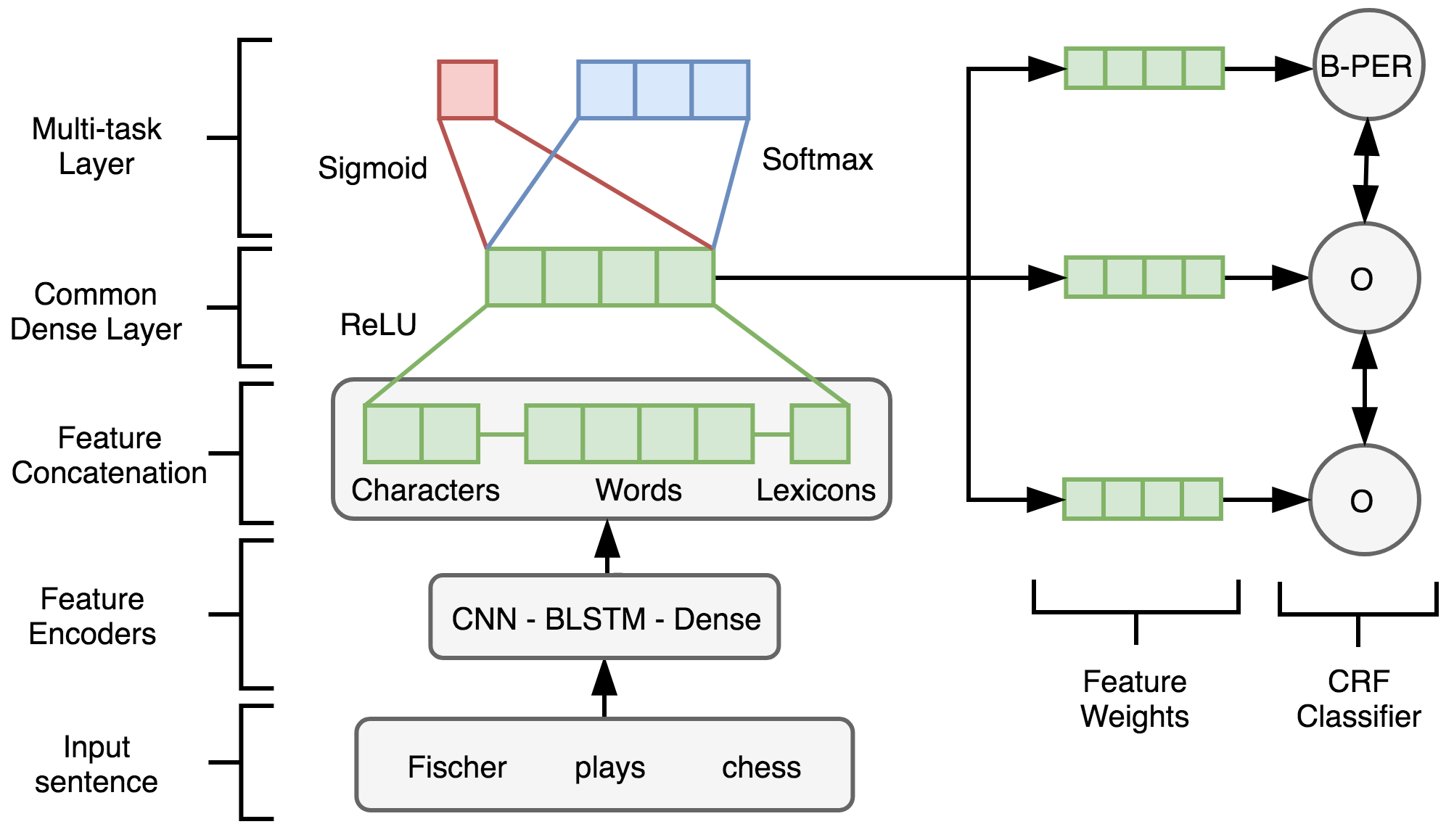}
\caption{ \small Overall system design. First, the system embeds a sentence into a high-dimensional space and uses CNN, BLSTM, and dense encoders to extract features. Then, it concatenates the resulting vectors of each encoder and performs multi-task. The top left single-node layer represents segmentation (red) while the top right three-node layer represents categorization (blue). Finally, a CRF classifier uses the weights of the common dense layer to perform a sequential classification. }
\label{fig:system_overview}
\end{figure}

The multi-task network predicts probabilities for each token in the input sentence individually. Thus, those individual probabilities do not account for sequential information. We exploit the sequential information by using a Conditional Random Fields\footnote{~Python CRF-Suite library: \url{https://github.com/scrapinghub/python-crfsuite}} classifier over those probabilities. This allows us to jointly predict the most likely sequence of labels for a given sentence instead of performing a word-by-word prediction. More specifically, we take the weights learned by the multi-task neural network and use them as features for the CRF classifier (see Figure \ref{fig:system_overview}). Taking weights from the common dense layer captures both of the segmentation and categorization features. 

\section{Experimental Settings} 

We preprocess all the datasets by replacing the URLs with the token \textless URL\textgreater~before performing any experiment. Additionally, we use half of development set as validation and the other half as evaluation.

Regarding the network hyper-parameters, in the case of the CNN, we set the kernel size to 3 on both convolutional layers. We also use the same number of filters on both layers: 64. Increasing the number of filters and the number of convolutional layers yields worse results, and it takes significantly more time. In the case of the BLSTM architecture, we add dropout layers before and after the Bidirectional LSTM layers with dropout rates of 0.5. The dropout layers allow the network to reduce overfitting \cite{Srivastava:2014:DSW:2627435.2670313}. We also tried using a batch normalization layer instead of dropouts, but the experiment yielded worse results. The training of the whole neural network is conducted using a batch size of 500 samples, and 150 epochs. Additionally, we compile the model using the AdaMax optimizer \cite{DBLP:journals/corr/KingmaB14}. Accuracy and F1-score are used as evaluation metrics.

For sequential inference, the CRF classifier uses L-BFGS as a training algorithm with L1 and L2 regularization. The penalties for L1 and L2 are $1.0$ and $1.0e^{-3}$, respectively.

\begin{table}
\resizebox{\columnwidth}{!}{%
\centering
\begin{tabular}{|l|r|r|r|}
\hline 
\bf Classes & 
\bf Precision (\%)& 
\bf Recall (\%) & 
\bf F1 (\%) \\ 
\hline\hline
corporation    & 35.71 & 29.41 & 32.26 \\
creative-work  & 60.00 &  5.26 &  9.68 \\
group          & 30.00 & 12.00 & 17.14 \\
location       & 65.71 & 56.10 & 60.53 \\
person         & 83.98 & 62.04 & 71.36 \\
product        & 39.29 & 15.71 & 22.45 \\
\hline
\hline
\bf Entity  & 72.16 & 43.30 & 54.12 \\
\bf Surface & 68.38 & 95.05 & 79.54 \\
\hline
\end{tabular}
}
\caption{\small \label{exp_dev_crf} This table shows the results from the CRF classifier at the class level. The classification is conducted using the development set as both validation and evaluation.}
\end{table}

\section{Results and Discussion}

We compare the results of the multi-task neural network itself and the CRF classifier on each of our experiments. The latter one always shows the best results, which emphasizes the importance of sequential information. The results of the CRF, using the development set, are in Table \ref{exp_dev_crf}.

Moreover, the addition of a secondary task allows the CRF to use more relevant features from the network improving its results from a F1-score of 52.42\% to 54.12\%. Our finding that a multi-task architecture is generally preferable over the single task  architecture is consistent with prior research~\cite{sogaard2016deep,collobert2008unified,attia-EtAl:2016:CogALex-V,maharjan-EtAl:2017:EACLlong}.

We also study the relevance of our features by performing multiple experiments with the same architecture and different combinations of features. For instance, removing gazetteers from the model drops the results from 54.12\% to 52.69\%. Similarly, removing POS tags gives worse results (51.12\%). Among many combinations, the feature set presented in Section \ref{feature_rep} yields the best results.

\begin{table}
\resizebox{\columnwidth}{!}{%
\centering
\begin{tabular}{|l|r|r|r|}
\hline 
\bf Classes & 
\bf Precision (\%) & 
\bf Recall (\%) & 
\bf F1 (\%) \\ 
\hline\hline
corporation    & 31.91 & 22.73 & 26.55 \\
creative-work  & 36.67 &  7.75 & 12.79 \\
group          & 41.79 & 16.97 & 24.14 \\
location       & 56.92 & 49.33 & 52.86 \\
person         & 70.72 & 50.12 & 58.66 \\
product        & 30.77 &  9.45 & 14.46 \\
\hline
\hline
\bf Entity	& 57.54 & 32.90 & 41.86 \\
\bf Surface & 56.31 & 31.31 & 40.24 \\
\hline
\end{tabular}
}
\caption{\small \label{test_results} This table shows the final results of our submission. The hardest class to predict for is \textit{creative-work}, while the easiest is \textit{person}.}
\end{table}

The final results of our submission to the WNUT-2017 shared task are shown in Table \ref{test_results}. Our approach obtains the best results for the \textit{person} and \textit{location} categories. It is less effective for \textit{corporation}, and the most difficult categories for our system are \textit{creative-work} and \textit{product}. Our intuition is that the latter two classes are the most difficult to predict for because they grow faster and have less restrictive patterns than the rest. For instance, products can have any type of letters or numbers in their names, or in the case of creative works, as many words as their titles can hold (e.g. name of movies, books, songs, etc.).

Regarding the shared-task metrics, our approach achieves a 41.86\% F1-score for entities and 40.24\% for surface forms. Table \ref{all-scores} shows that our system yields similar results to the other participants on both metrics. In general, the final scores are low which states the difficulty of the task and that the problem is far from being solved.

\begin{table}
 \small
\centering
\begin{tabular}{|l|r|r|}
\hline 
\bf Participants & 
\bf F1 - E (\%) & 
\bf F1 - SF (\%) \\ 
\hline
\hline
MIC-CIS         & 37.06 & 34.25 \\ 
Arcada          & 39.98 & 37.77 \\ 
Drexel-CCI      & 26.30 & 25.26 \\ 
SJTU-Adapt      & 40.42 & 37.62 \\ 
FLYTXT          & 38.35 & 36.31 \\ 
SpinningBytes   & 40.78 & 39.33 \\ 
\bf{UH-RiTUAL}  & \bf{41.86} & \bf{40.24} \\
\hline
\end{tabular}
\caption{\small  \label{all-scores} The scores of all the participants in the WNUT-2017 shared task. The metrics of the shared task are entity and surface form F1-scores. Our results are highlighted. }
\end{table}

\section{Error Analysis}

By evaluating the errors made by the CRF classifier, we find that the NE boundaries are a problem. For instance, when a NE is preceded by an article starting with a capitalized letter, the model includes the article as if it were part of the NE. This behavior may be caused by the capitalization features captured by the CNN network. Similarly, if a NE is followed by a conjunction and another NE, the classifier tends to join both NEs as if the conjunction were part of a single unified entity. Another common problem shown by the classifier is that fully-capitalized NEs are disregarded most of the time. This pattern may be related to the switch of domains in the training and testing phases. For instance, some Twitter informal abbreviations\footnote{~E.g. \textit{LOL} is an informal social media expression that stands for \textit{Laughing Out Loud}, which is not an NE.} may appear fully-capitalized but they do not represent NEs, whereas in Reddit and Stack Overflow fully-capitalized words are more likely to describe NEs.

\section{Conclusion}

We show that our multi-task neural network is capable of extracting relevant features from noisy user-generated text. We also show that a CRF classifier can boost the neural network results because it uses the whole sentence to predict the most likely set of labels. Additionally, our approach emphasizes the importance of POS tags in conjunction with gazetteers for NER tasks. Twitter word embeddings and orthographic character embeddings are also relevant for the task.

Finally, our ongoing work aims at improving these results by getting a better understanding of the strengths and weaknesses of our model. We also plan to evaluate the current system in related tasks where noise and emerging NEs are prevalent.

\bibliography{emnlp2017}
\bibliographystyle{emnlp_natbib}

\end{document}